%% file: main.tex
\pdfoutput=1

\documentclass[11pt]{article}

\usepackage{acl}

\usepackage{times}
\usepackage{latexsym}

\usepackage[T1]{fontenc}

\usepackage[utf8]{inputenc}

\usepackage{microtype}

\input{math_commands.tex}

\usepackage{amsmath}
\usepackage{multirow}
\usepackage{xcolor}

\usepackage{xcolor,colortbl}
\definecolor{lightblue}{rgb}{0.68, 0.85, 0.9}
\definecolor{lavender}{rgb}{0.9, 0.9, 0.98}
\definecolor{lightyellow}{rgb}{1.0, 1.0, 0.88}
\definecolor{magicmint}{rgb}{0.67, 0.94, 0.82}
\definecolor{palepink}{rgb}{0.98, 0.85, 0.87}
\definecolor{bubbles}{rgb}{0.91, 1.0, 1.0}

\usepackage{microtype}
\usepackage{graphicx}
\usepackage{subfigure}
\usepackage{amssymb}
\usepackage{CJKutf8}
\usepackage{algorithm}
\usepackage{algorithmic}
\usepackage{enumitem}
\algsetup{linenosize=\small}
\usepackage{wrapfig}
\usepackage{cleveref}
\usepackage{multicol}
\usepackage{booktabs}
\usepackage{xspace}
\usepackage{adjustbox}
\usepackage{subfiles}
\usepackage{latexsym}

\usepackage{placeins}
\usepackage[french,vietnamese,main=english]{babel}

\newcommand{\cchar}[1]{\begin{CJK*}{UTF8}{gkai}#1\end{CJK*}}

\newcommand*{\affaddr}[1]{#1} 
\newcommand*{\affmark}[1][*]{\textsuperscript{#1}}
\newcommand*{\email}[1]{\text{#1}}
\renewcommand{\secref}[1]{\S\ref{#1}}
\newcommand{\name}{\mbox{GlobalWoZ}\xspace}

\newcommand{\notationa}{Zero-Shot (E\&E)\xspace}
\newcommand{\notationb}{Translate-Train\xspace}
\newcommand{\notationc}{SUC\xspace}
\newcommand{\notationd}{BBUC\xspace}
\newcommand{\notatione}{MBUC\xspace}
\newcommand{\notationf}{MMUC\xspace}

\newcommand{\notationC}{\textbf{SUC}\xspace}

\newcommand{\usecasea}{F\&F\xspace}
\newcommand{\usecaseb}{F\&E\xspace}
\newcommand{\usecasec}{E\&F\xspace}
\newcommand{\usecased}{E\&E\xspace}

\usepackage{bbding}
\usepackage{multirow}

\title{GlobalWoZ: Globalizing MultiWoZ to Develop Multilingual Task-Oriented Dialogue Systems}

\author{Bosheng Ding\thanks{\; Bosheng Ding is under the Joint PhD Program between Alibaba and Nanyang Technological University.}\affmark[~~1,2]~~ Junjie Hu\affmark[3]~~ Lidong Bing\affmark[1]~~   \\ \textbf{Sharifah Mahani Aljunied\affmark[1]~~
Shafiq Joty\affmark[2]~~ Luo Si\affmark[1]~~ Chunyan Miao\affmark[2]}\\
\affaddr{\affmark[1]DAMO Academy, Alibaba Group}
\affaddr{\affmark[2]Nanyang Technological University, Singapore}\\
\affaddr{\affmark[3]University of Wisconsin-Madison}\\
\email{\small{\tt \{bosheng.ding, l.bing, mahani.aljunied, luo.si\}@alibaba-inc.com}}\\
\email{\small{\tt junjie.hu@wisc.edu,}}
\email{\small{\tt\{srjoty, ascymiao\}@ntu.edu.sg}}}

\begin{document}
\maketitle
\begin{abstract}
\input{00_abstract}

\end{abstract}

\input{01_introduction}
\input{03_methodology}

\input{04_experiments}
\input{05_baselines}
\input{06_results}
\input{07_analysis}

\input{02_related_work}
\input{08_conclusion}

\bibliography{anthology_aclrr}
\bibliographystyle{acl_natbib}

\clearpage
\newpage
\onecolumn
\appendix
\input{09_appendix}

\end{document}

%% file: math_commands.tex
\usepackage{amsmath,amsfonts,bm}

\def\secref#1{section~\ref{#1}}

\def\eqref#1{equation~\ref{#1}}

\def\1{\bm{1}}

\DeclareMathAlphabet{\mathsfit}{\encodingdefault}{\sfdefault}{m}{sl}
\SetMathAlphabet{\mathsfit}{bold}{\encodingdefault}{\sfdefault}{bx}{n}



%% file: 00_abstract.tex



Over the last few years, there has been a move towards data curation for multilingual task-oriented dialogue (ToD) systems that can serve people speaking different languages. However, existing multilingual ToD datasets either have a limited coverage of languages due to the high cost of data curation, or ignore the fact that dialogue entities barely exist in countries speaking these languages. To tackle these limitations, we introduce a novel data curation method that generates \textbf{\name} --- a large-scale multilingual ToD dataset globalized from an English ToD dataset for three unexplored use cases of  multilingual ToD systems. Our method is based on translating dialogue templates and filling them with local entities in the target-language countries. Besides, we extend the coverage of target languages to 20 languages. We will release our dataset and a set of strong baselines to encourage research on multilingual ToD systems for real use cases.\footnote{Our code is available at \url{https://ntunlpsg.github.io/project/globalwoz/}.}



%% file: 01_introduction.tex
\section{Introduction}
\label{sec:introduction}

\begin{figure*}[t!]
\centering
\includegraphics[scale=0.66]{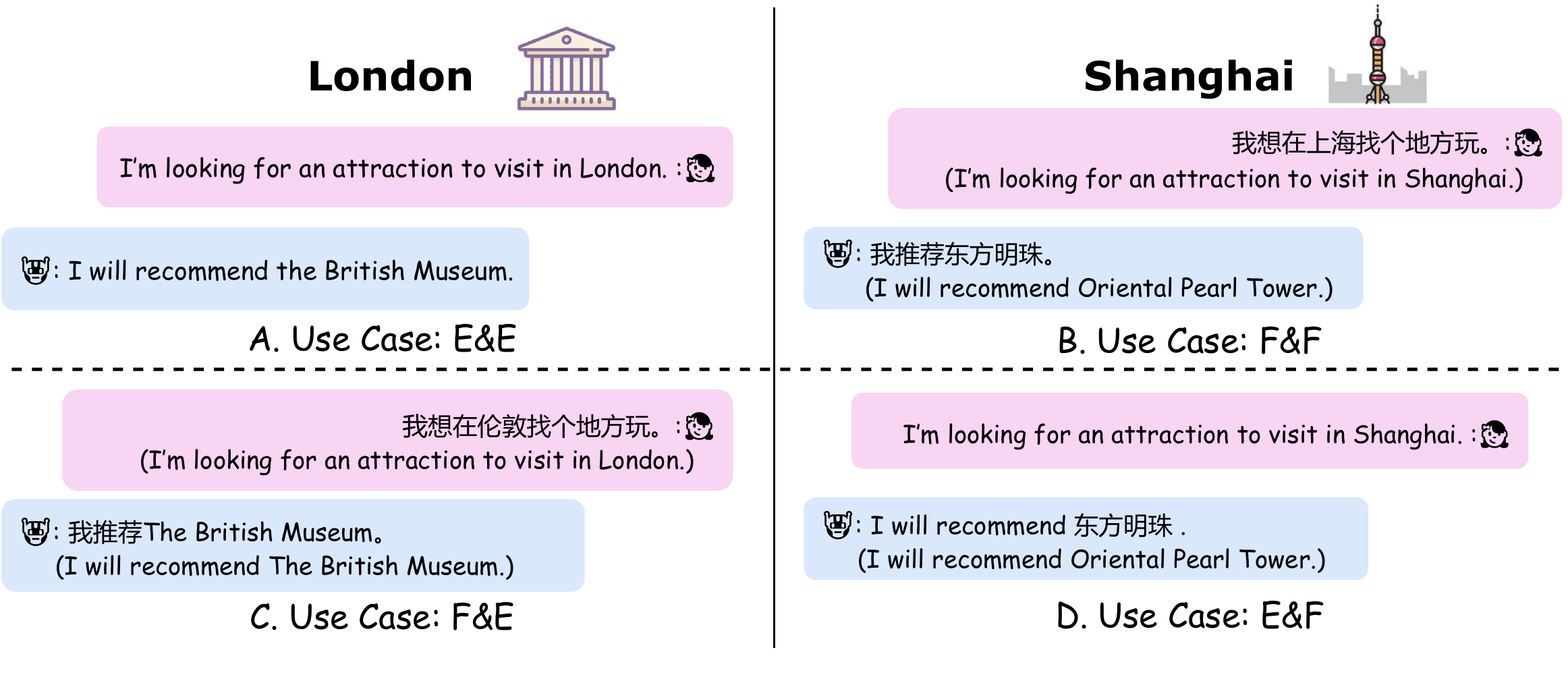}
\caption{Examples of four use cases for multilingual ToD systems: A. Use Case \usecased: A English speaker travels to a country of English.  B. Use Case \usecasea: A foreign language speaker travels to a country of the foreign language.  C. Use Case \usecaseb: A foreign language speaker travels to a country of English. D. Use Case \usecasec: A English speaker travels to a country of a foreign language.
}
\label{fig:illustration}
\end{figure*}

One of the fundamental objectives in pursuit of artificial intelligence is to enable machines with the ability to intelligently communicate with human in natural languages, with one of the widely-heralded applications being the task-oriented dialogue (ToD) systems \citep{Gupta2006TheAS,Bohus2009TheRD}. Recently, ToD systems have been successfully deployed to assist users with accomplishing certain domain-specific tasks such as hotel booking, alarm setting or weather query \citep{eric-etal-2017-key,wu-etal-2019-transferable,lin-etal-2020-mintl,zhang2020recent}, thanks to the joint advent of neural networks and availability of domain-specific data. However, most existing ToD systems are predominately built for English, limiting their service for \textit{all} of the world's citizens.  The reason of this limitation lies in the stark lack of high-quality multilingual ToD datasets due to the high expense and challenges of human annotation~\citep{razumovskaia2021crossing}. 

One solution to this is annotating conversations in other languages from scratch, e.g., CrossWoZ \cite{zhu2020crosswoz} and BiToD \cite{lin2021bitod}. However, these methods involve expensive human efforts for dialogue collection in the other languages, resulting in a limited language/domain coverage. The other major line of work focused on translating an existing English ToD dataset into target languages by professional human translators \cite{Upadhyay2018AlmostZC,schuster2018cross,van-der-goot-etal-2021-masked,li-etal-2021-mtop}. Despite the increasing language coverage, these methods simply translated English named entities (e.g., location, restaurant name) into the target languages, while ignored the fact that these entities barely exist in countries speaking these  languages. This hinders a trained ToD system from supporting the real use cases where a user looks for local entities in a target-language country. For example in Figure~\ref{fig:illustration}, a user may look for the British Museum when traveling to London (A.), while look for the Oriental Pearl Tower when traveling to Shanghai (B.). 

In addition, prior studies~\cite{cheng1989code,kim2006reasons} have shown that code-switching phenomena frequently occurs in a dialogue when a speaker cannot express an entity immediately and has to alternate between two languages to convey information more accurately. Such phenomena could be ubiquitous during the cross-lingual and cross-country task-oriented conversations. One of the reasons for code-switching is that there are no exact translations for many local entities in the other languages. Even though we have the translations, they are rarely used by local people. For example in Figure~\ref{fig:illustration} (C.), after obtaining the recommendation from a ToD system, a Chinese speaker traveling to London would rather use the English entity ``British Museum'' than its Chinese translation to search online or ask local people. To verify this code-switching phenomena, we have also conducted a case study (\secref{sec:validation}) which shows that searching the information about translated entities online yields a much higher failure rate than searching them in their original languages. 
Motivated by these observations, we define \textit{three unexplored use cases}\footnote{See comparisons of these use cases in Appendix~\ref{appendix:comparison}} of multilingual ToD where a foreign-language speaker uses ToD in the foreign-language country (\textbf{\usecasea}) or an English country (\textbf{\usecaseb}), and an English speaker uses ToD in a foreign-language country (\textbf{\usecasec}). These use cases are different from the traditional \textbf{\usecased} use case where an English speaker uses ToD in an English-speaking country.

To bridge the aforementioned gap between existing data curation methods and the real use cases, we propose a novel data curation method that \textit{globalizes} an existing multi-domain ToD dataset beyond English for the three unexplored use cases. Specifically, building on top of MultiWoZ~\cite{budzianowski-etal-2018-multiwoz} --- an English ToD dataset for dialogue state tracking (DST), we create \name, a new multilingual ToD dataset in three new target-languages via machine translation and crawled ontologies in the target-language countries. 

Our method only requires minor human efforts to post-edit a few hundred machine-translated dialogue templates in the target languages for evaluation. 
Besides, as cross-lingual transfer via pre-trained multilingual models~\cite{devlin-etal-2019-bert,conneau-etal-2020-unsupervised, liu2020multilingual, xue-etal-2021-mt5} has proven effective in many cross-lingual tasks, we further investigate another question: \textit{How do these multilingual models trained on the English ToD dataset transfer knowledge to our globalized dataset?} To answer this question, we prepare a few baselines by evaluating popular ToD systems on our created test datasets in a \textit{zero-shot} cross-lingual transfer setting as well as a \textit{few-shot} setting. 


Our contributions include the following: 
\begin{itemize} [leftmargin=10pt]\itemsep-0.2em
    \item To the best of our knowledge, we provide the first step towards analyzing three unexplored use cases for multilingual ToD systems.
    \item We propose a cost-effective method that creates a new multilingual ToD dataset from an existing English dataset. Our dataset consists of high-quality test sets which are first translated by machines and then post-edited by professional translators in three target languages (Chinese, Spanish and Indonesian). We also leverage machine translation to extend the language coverage of test data to another 17 target languages.
    \item Our experiments show that current multilingual systems and translate-train methods fail in zero-shot cross-lingual transfer on the dialogue state tracking task. To tackle this problem, we propose several data augmentation methods to train strong baseline models in both zero-shot and few-shot cross-lingual transfer settings. 
\end{itemize}


%% file: 03_methodology.tex
\section{Data Curation Methodology}
\label{sec:methodology}



\begin{figure*}[t!]
\centering
\includegraphics[scale=0.48]{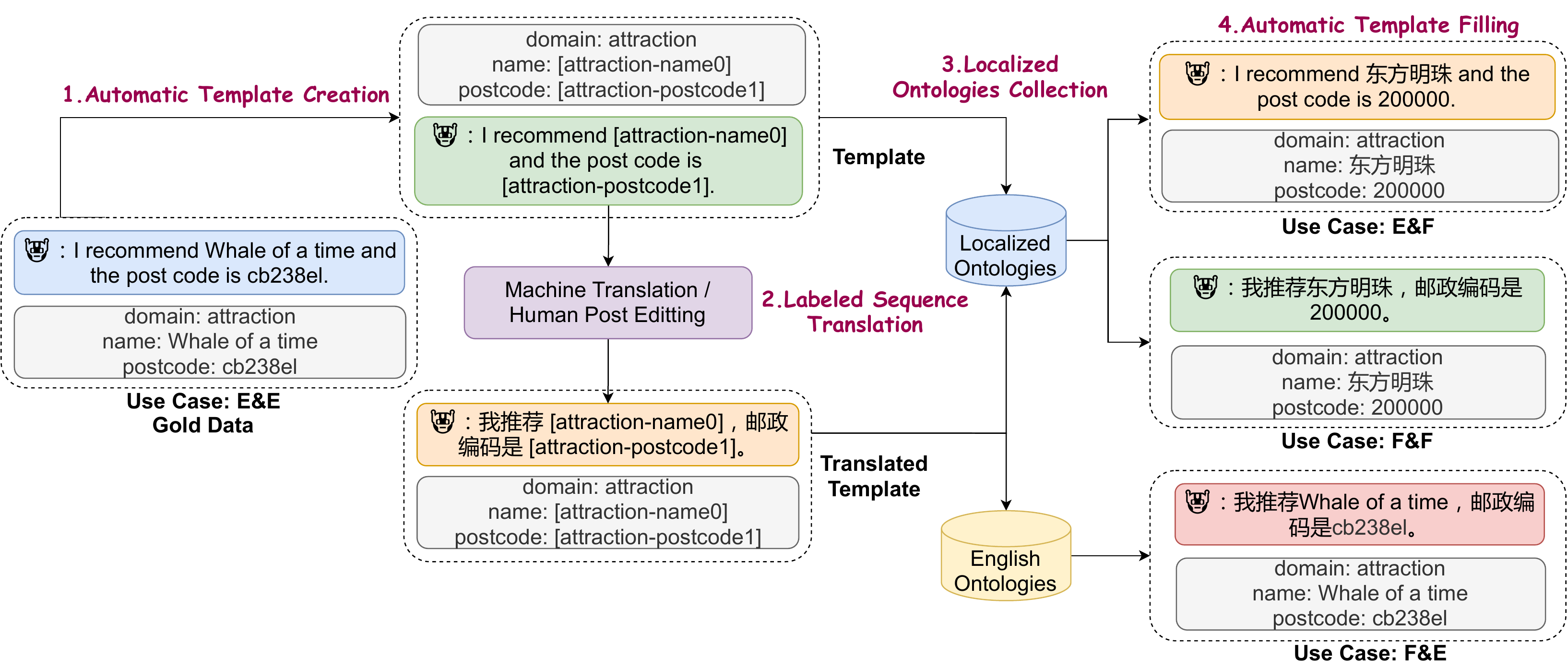}
\caption{Illustration of our proposed pipeline: 1. Automatic Template Creation 2. Labeled Sequence Translation 3. Localized Ontologies Collection 4. Automatic Template Filling
}
\label{fig:flow}
\end{figure*}

In order to globalize an existing English ToD dataset for the three aforementioned use cases, we propose an approach consisting of four steps as shown in Figure~\ref{fig:flow}: (1) we first extract dialogue templates from the English ToD dataset by replacing English-specific entities with a set of general-purpose placeholders (\secref{sec:template}); (2) we then translate the templates to a target language for both training and test data, with one key distinction that we only post-edit the test data by professional translators to ensure the data quality for evaluation (\secref{sec:translate}); (3) next, we collect ontologies \cite{kiefer2021vonda} containing the definitions of dialogue acts, local entities and their attributes in the target-language countries (\secref{sec:ontology}); (4) finally, we tailor the translated templates by automatically substituting the placeholders with entities in the extracted ontologies to construct data for the three use cases (\secref{sec:use-case}). 



\subsection{Automatic Template Creation} \label{sec:template}

We start with MultiWoZ 2.2 \cite{zang-etal-2020-multiwoz} -- a high-quality multi-domain English ToD dataset with more accurate human annotations compared to its predecessors MultiWoZ 2.0 \cite{budzianowski-etal-2018-multiwoz} and MultiWoz 2.1 \cite{eric-EtAl:2020:LREC}. For the sake of reducing human efforts for collecting ToD context in the target languages, we re-use the ToD context written by human in MultiWoZ as the dialogue templates. Specifically as shown in Figure~\ref{fig:flow}, we replace the English entities in MultiWoz by a set of general-purpose placeholders such as \texttt{[attraction-name0]} and \texttt{[attraction-postcode1]}, where each placeholder contains the entity's domain, attribute and ID. To do so, we first build a dictionary with entity-placeholder pairs by parsing the annotations of all dialogues. For example, from a dialogue text ---\textit{``I recommend Whale of a time and the post code is cb238el.''}, we obtain two entity-placeholder pairs from its human annotations, i.e., (\textit{Whale of a time}, \texttt{[attraction-name0]}) and (\textit{cb238el}, \texttt{[attraction-postcode1]}). Next, we identify entities in the dialogue by their word index from the human annotations, replace them with their placeholders in the dictionary, and finally obtain dialogue templates with placeholders. Notably, we skip the entities with their attributes of \texttt{[choice]} and \texttt{[ref]} that represent the number of choices and booking reference number, as these attributes could be used globally.



\subsection{Labeled Sequence Translation}\label{sec:translate}
Following \citet{liu-etal-2021-mulda} that translates sentences with placeholders, we use a machine translation system\footnote{We use Google Translate (\url{https://cloud.google.com/translate}), an off-the-shelf MT system.} to translate dialogue templates with our designed placeholders. As we observe, a placeholder containing an entity domain, attribute and ID (e.g., \texttt{attraction-name0}) is useful to provide contextually meaningful information to the translation system, thus usually resulting in a high-quality translation with the placeholder unchanged \footnote{Appendix~\ref{appendix:sequence_translation} has an example of label sequence translation.}. This also enables us to easily locate the placeholders in the translation output and replace them with new entities in the target language.

To build a high-quality test set for evaluation, we further hire professional translators to post-edit a few hundred machine-translated templates, which produces natural and coherent sentences in the target languages.\footnote{Appendix~\ref{appendix:bleu} shows the bleu scores between MT test data and MTPE test data.} With the goal of selecting representative test templates for post-editing, we first calculate the frequency of all the 4-gram combinations in the MultiWoZ data, and then score each dialogue in the test set by the sum of the frequency of all the 4-gram combinations in the dialogue divided by the dialogue's word length. We use this scoring function to estimate the representiveness of a dialogue in the original dataset. Finally, we select the top 500 high-scoring dialogues in the test set for
post-editing.\footnote{Appendix~\ref{appendix:test_dist} shows the English test data distribution.} 
We also use the same procedure to create a small high-quality training set for few-shot cross-lingual transfer setting.

\subsection{Collection of Local Ontology} \label{sec:ontology}
Meanwhile, we crawl the attribute information of local entities in three cities from public websites (e.g., tripadvisor.com, booking.com) to create three ontologies for the three corresponding target languages respectively. As shown in Table~\ref{tb:selected_languages} in Appendix~\ref{appendix:sel_lang}, we select Barcelona for Spanish (an Indo-European language), Shanghai for Mandarin (a Sino-Tibetan language) and Jakarta for Indonesian (an Austronesian language), which cover a set of typologically different language families. 

Given a translated dialogue template, we can easily sample a random set of entities for a domain of interest from a crawled ontology and assign the entities to the template's placeholders to obtain a new dialogue in the target language. Repeating this procedure on each dialogue template, we can easily build a high-quality labeled dataset in the target language. Table~\ref{tb:stats_ontology_17} in Appendix~\ref{appendix:stat_ontology} shows the statistics of our collected entities in the target languages compared with the English data. The number of our collected entities are either larger than or equal to those in the English data except for the ``train'' domain; we collected the information about only 100 ``trains'' for each languages due to the complexity in collecting relevant information. 

\subsection{Template Filling for Three Use Cases} \label{sec:use-case}
After the above steps, we assign entities in a target language to the translated templates in the same target language for the \usecasea case, while assigning target-language entities to the English (source-language) templates for the \usecaseb case. As for the \usecasec case, we keep the original English context by skipping the translation step and replace the placeholders with local entities in the target language (see Figure \ref{fig:flow} for examples). 

To sum up, our proposed method has three key properties: (1) our method is \textit{cost-effective} as we only require a limited amount of post-editing efforts for a test set when compared to the expensive crowd-sourced efforts from the other studies; (2) we can easily sample entities from an ontology to create \textit{large-scale machine-translated data} as a way of data augmentation for training; 
(3) our method is \textit{flexible} to update entities in a ToD system whenever an update of ontology is available, e.g., extension of new entities. We refer the readers to Table~\ref{tb:stats_globalwoz} for the data statistics of \name and Figure~\ref{fig:example} for dialogue examples in the appendix. 



%% file: 04_experiments.tex
\section{Task \& Settings}
\label{sec:tasks}

\subsection{Dialogue State Tracking}
Our experiments focus on the dialogue state tracking (DST), one of the fundamental components in a ToD system that predicts the goals of a user query in multi-turn conversations. We follow the setup in MultiWoZ~\cite{budzianowski-etal-2018-multiwoz} to evaluate ToD systems for DST by the joint goal accuracy which measures the percentage of correctly predicting all goals in a multi-turn conversation.

\subsection{Experimental Settings}

\noindent\textbf{Zero-Shot Cross-lingual Transfer:} Unlike prior studies that annotate a full set of high-quality training data for a target language, we investigate the \textit{zero-shot cross-lingual transfer} setting where we have access to only a high-quality human-annotated English ToD data (referred to as gold standard data hereafter). In addition, we assume that we have access to a machine translation system that translates from English to the target language. We investigate this setting to evaluate how a multilingual ToD system transfers knowledge from a high-resource source language to a low-resource target language.

\noindent\textbf{Few-Shot Cross-lingual Transfer:} We also investigate few-shot cross-lingual transfer, a more practical setting where we are given a small budget to annotate ToD data for training. Specifically, we include a small set (100 dialogues) of high-quality training data post-edited by professional translators (\secref{sec:translate}) in a target language, and evaluate the efficiency of a multilingual ToD on learning from a few target-language training examples.

%% file: 05_baselines.tex
\section{Proposed Baselines}
\label{sec:baselines}
 
We prepare a base model for \name in the zero-shot and few-shot cross-lingual transfer settings. We select Transformer-DST \citep{zeng2020jointly} as our base model as it is one of the state-of-the-art models on both MultiWoZ 2.0 and MultiWoZ 2.1\footnote{According to the leaderboards of Multi-domain Dialogue State Tracking on MultiWoZ 2.0 and MultiWoZ 2.1 on paperwithcode.com as of 11/15/2021.}. In our paper, we replace its BERT encoder with an mBERT encoder \citep{devlin-etal-2019-bert} for our base model and propose a series of training methods for \name. As detailed below, we propose several data augmentation baselines that create different training and validation data for training a base model. 
Note that all the proposed baselines are model agnostic and the base model can be easily substituted with other popular models \cite{heck-etal-2020-trippy,lin-etal-2020-mintl}. For each baseline, we first train a base model on its training data for 20 epochs and use its validation set to select the best model during training. Finally we evaluate the best model of each baseline on the same test set from \name. We will release \name and our pre-trained models to encourage faster adaptation to future research. 
We refer the readers to Table~\ref{tb:Methods_data} and Table~\ref{tb:Methods_data_access} in Appendix~\ref{appendix:summary_baselines} 
while reading the subsequent methods for a better understanding. 

\subsection{Pure Zero-Shot (\usecased)}
We train a base model on the gold standard English data (\usecased) and directly apply the learned model to the test data of the three use cases in \name. With this method, we simulate the condition of having labeled data only in the source language for training, and evaluate how the model transfers knowledge from English to the three use cases. 
We use \textbf{\notationa} to denote this method.

\subsection{Translate-Train}
We use our data curation method (\secref{sec:methodology}) to translate the templates by an MT system but replace the placeholders in the translated templates with machine-translated entities to create a set of pseudo-labeled training data. Next, we train a base model on the translated training data without local entities, and evaluate the model on the three use cases. 
We denote this method as \textbf{\notationb}.

\subsection{Single-Use-Case Training} 
By skipping the human post-editing step in our data curation method (\secref{sec:methodology}), we leverage a machine translation system to automatically create a large set of pseudo-labeled training data with local entities for the three use cases. In the \usecasea case, we translate the English templates by the MT system and replace the placeholders in the translated templates with foreign-language entities to create a training dataset. In the \usecaseb case, we replace the placeholders in the translated templates with the original English entities to create a code-switched training dataset. In the \usecasec case, we use the original English templates and replace the placeholders in the English templates with foreign-language entities to create a code-switch training dataset. With this data augmentation method, we can train a base model on each pseudo-labeled training dataset created for each use case. 
We denote this method as \textbf{\notationc} (Single-Use-Case).

\subsection{Bi-/Multi-lingual Bi-Use-Case Training}
We investigate the performance of combining the existing English data and the pseudo-labeled training data created for one of the three use cases (i.e., \usecasea, \usecaseb, \usecasec), one at a time, to do bi-use-case training. In the bilingual training, we only combine the gold English data (\usecased) with the pseudo-labeled training data in one target language in one use case for joint training. We denote this method as \textbf{\notationd} (Bilingual Bi-Use-Case). In the multilingual training, we combine gold English data (\usecased) and pseudo-labeled training data in all languages in one use case for joint training.  We denote this method as \textbf{\notatione} (Multilingual Bi-Use-Case).

\subsection{Multilingual Multi-Use-Case Training}
We also propose to combine the existing English data (\usecased) and all the pseudo-labeled training data in all target languages for all the use cases (\usecasea, \usecaseb, \usecasec). We then train a single model on this combined multilingual training dataset and evaluate the model on test data in all target languages for all three use cases . 
We denote this method as \textbf{\notationf} (Multilingual Multi-Use-Case).

%% file: 06_results.tex
\section{Experiment Results}
\label{sec:experiments}
In this section, we show the results of all methods in the zero-shot~(\secref{sec:exp_zero_shot}) and few-shot~(\secref{sec:exp_few_shot}) settings.

\subsection{Zero-shot Cross-lingual Transfer} \label{sec:exp_zero_shot}
\subsubsection{Use Case \usecasea, \usecaseb and \usecasec}
Table~\ref{tb:Use_case_result} reports the joint goal accuracy of all proposed methods on the three different sets of test data in the \usecasea, \usecaseb, and \usecasec use cases\footnote{Appendix~\ref{appendix:EE} reports the results in the \usecased use case.}. 
Both \notationa and \notationb struggle, achieving average accuracy of less than 10 in all use cases. Despite its poor performance, \notationa works much better in \usecaseb than \usecasea, while its results in \usecasea and \usecasec are comparable, indicating that a zero-shot model trained in \usecased can transfer knowledge about local English entities more effectively than knowledge about English context in downstream use cases. 
Besides, we also find that \notationa performs better on the Spanish or Indonesian context than the Chinese context in \usecaseb. One possible reason is that English is closer to the other Latin-script languages (Spanish and Indonesian) than Chinese. 

Our proposed data augmentation methods (\notationc, \notationd, \notatione) perform much better than non-adapted methods (\notationa and \notationb) that do not leverage any local entities for training. In particular, it is worth noting that even though \notationb and \notationc both do training on foreign-language entities in \usecasea and \usecasec, there is a huge gap between these two methods, since \notationb has only access to the machine-translated entities rather than the real local entities used by \notationc. This huge performance gaps not only show that \notationb is not an effective method in practical use cases but also prove that having access to local entities is a key to building a multilingual ToD system for practical usage.

Comparing our data augmentation methods \notationc and \notationd, we find that the base model can benefit from training on additional English data (E\&E), especially yielding a clear improvement of up to 5.58 average accuracy points in \usecaseb. Moreover, when we increase the number of languages in the bi-use-case data augmentations (i.e., \notatione), we observe an improvement of around 1 average accuracy points in all three use cases w.r.t. \notationd. These observations encourage a potential future direction that explores better data augmentation methods to create high-quality pseudo-training data.  

\begin{table}[t!]
\centering
\resizebox{\columnwidth}{!}{%
\begin{tabular}{llcccc}
\toprule
\textbf{Case} & \textbf{Methods}          & \textbf{zh} & \textbf{es}  & \textbf{id} &\textbf{avg}\\
\midrule
\multirow{5}{*}{\usecasea}&\notationa          & 1.22  & 1.38  & 1.26  & 1.28  \\
& \notationb          & 2.61  & 2.59  & 5.74  & 3.65  \\
& \notationc (\usecasea) & 36.97 & 24.66 & 25.26 & 28.96 \\
& \notationd (\usecased+ \usecasea)      & 37.32 & 25.52 & 26.39 & 29.74 \\
& \notatione (\usecased+ \usecasea)      & \textbf{38.01} & \textbf{26.03} & \textbf{28.22} & \textbf{30.76} \\
\midrule
\multirow{5}{*}{\usecaseb}&\notationa             & 6.92           & 11.34          & 9.09           & 9.12  \\
&\notationb        & 2.28           & 4.97           & 4.67           & 3.97  \\
&\notationc (\usecaseb)            & 56.28          & 41.94          & 47.93          & 48.71 \\
&\notationd (\usecased+ \usecaseb)             & 59.87          & 48.20          & 54.79          & 54.29 \\
&\notatione (\usecased+ \usecaseb)   & \textbf{60.37} & \textbf{53.56} & \textbf{54.93} & \textbf{56.28} \\
\midrule
\multirow{5}{*}{\usecasec}&\notationa        & 1.69           & 1.81           & 1.82           & 1.77           \\
& \notationb         & 1.39           & 1.76           & 1.86           & 1.67           \\
& \notationc (\usecasec)        & 38.56          & 28.00          & 43.82          & 36.79          \\
& \notationd (\usecased+ \usecasec)      & 39.87          & 27.29          & 45.48          & 37.54          \\
& \notatione (\usecased+ \usecasec)      & \textbf{40.20} & \textbf{29.22} & \textbf{47.06}          & \textbf{38.83} \\
\bottomrule
\end{tabular}
}
\caption{Zero-shot cross-lingual accuracy on DST over three target languages in three use cases.}
\label{tb:Use_case_result}
\end{table}

\subsubsection{One Model for All}
\begin{figure}[t!]
\centering
\includegraphics[scale=0.5]{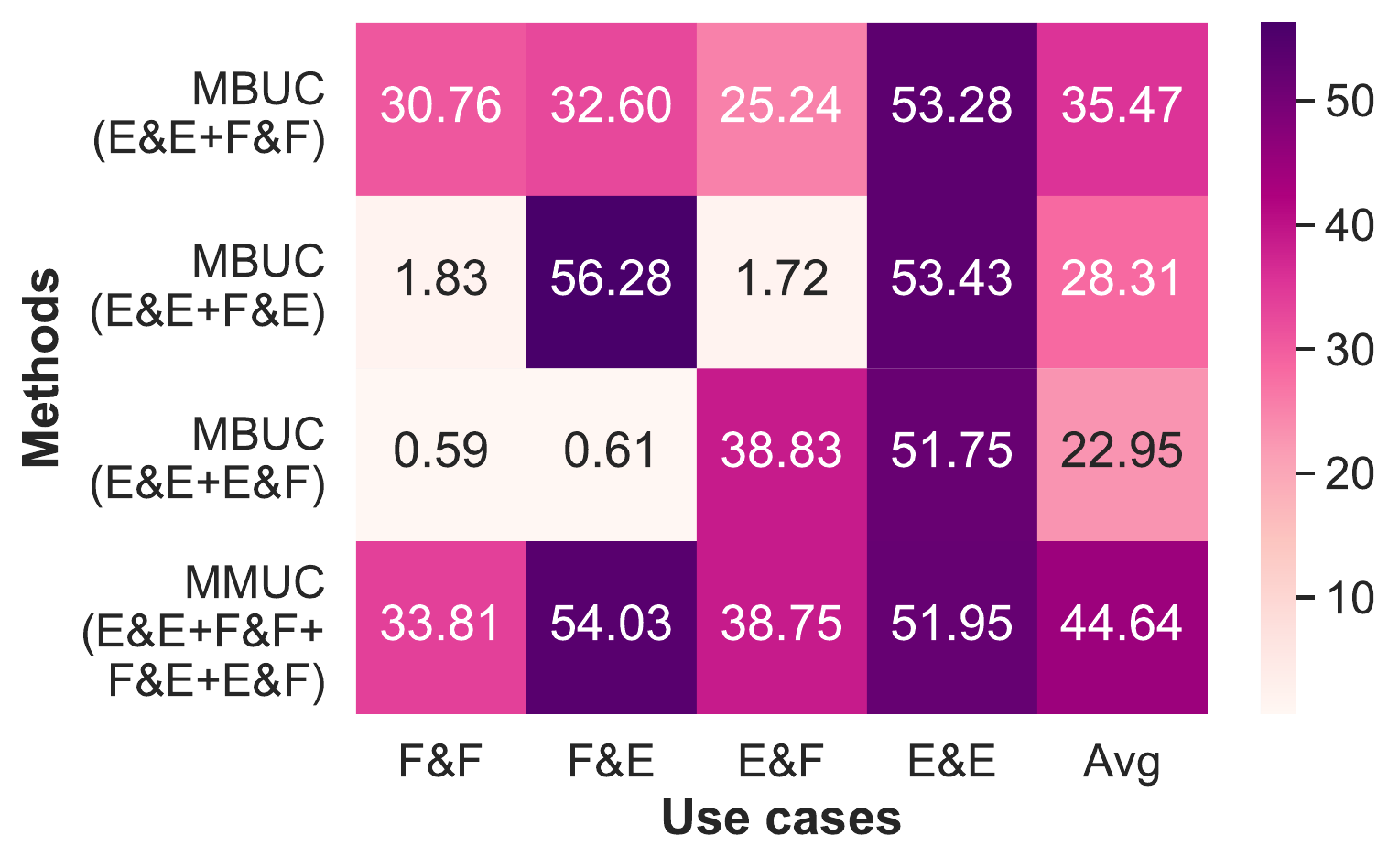}
\caption{Performance of \notationf vs \notatione on the test data of the four use cases, \usecasea, \usecaseb, \usecasec and  \usecased.
}
\label{fig:heatmap}
\end{figure}

Notice that we can train a single model by \notationf for all use cases rather than training separate models, one for each use case. In Figure~\ref{fig:heatmap}, we compare \notationf and \notatione (rows) on the test data in the four use cases (columns). Although \notationf may not achieve the best results in each use case, it achieves the best average result over the four use cases, indicating the potential of using one model to simultaneously handle all the four use cases. 


\subsection{Few-shot Cross-lingual Transfer}\label{sec:exp_few_shot}
\begin{figure}[t!]
\centering
\includegraphics[width=\columnwidth]{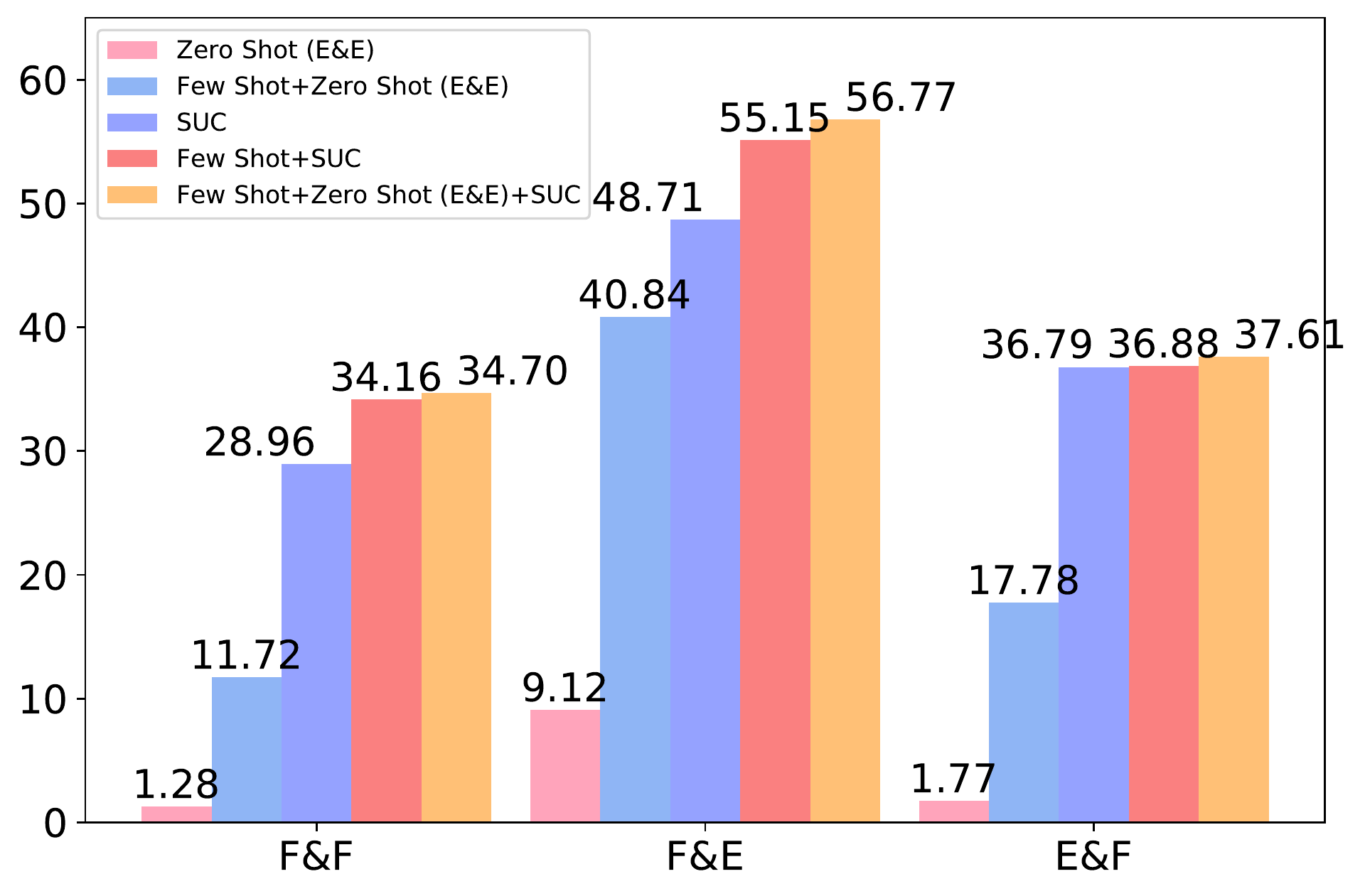}
\caption{Few-shot cross-lingual average joint accuracy on DST over three target languages in three use cases. 
}
\label{fig:few_shot_result}
\end{figure}

In few-shot experiments, we use the same scoring function based on frequency of all 4-gram combinations (\secref{sec:translate}) to select 100 additional dialogues from train set for human-post editing, and create high-quality training data for each of the three use cases. To avoid overfitting on this small few-shot dataset, we
combine the few-shot data with the existing English data for training a base model (Few-Shot+\notationa). Next, we also investigate a model trained with additional synthetic data created by our proposed \notationc.
In Figure~\ref{fig:few_shot_result}, we find that our proposed \notationc without additional few-shot data has already outperformed the model trained with few-shot data and English data (Few-shot + \notationa), indicating that the model benefit more from a large amount of pseudo-labeled data than a small set of human-labeled data. 
If we combine the data created by \notationc with the few-shot data or with both few-shot and English data to train the model, we observe improvements over \notationc, especially with a clear gain of 8.06 accuracy points in \usecaseb. We refer the readers to Table~\ref{tb:few_shot_breakdown} in the appendix for detailed scores in all target languages.


%% file: 07_analysis.tex
\section{Discussion}
\label{sec:analysis}

\subsection{Motivation for Code-Switched Use Cases}
\label{sec:validation}

One key research question is to validate whether code-switched use cases with local entities (i.e., \usecaseb, \usecasec) are practically more useful for information seeking. 
To answer this question, we compare the failure rate of using local entities and machine-translated entities in information search, which is a proxy to the efficiency of using these two types of entities in conversations. We first randomly select 100 entities (33 attractions, 33 hotels and 34 restaurants) of Cambridge, Shanghai, Barcelona and Jakarta.  We translate the English entities into Mandarin, Spanish and Indonesian and the foreign-language entities into English via Google Translate. We then manually search the translated entities on Google to check whether we can find the right information of the original entities. Notice that the failure of the above verification partially come from the translation error made by Google Translate, or the search failure due to the fact that this entity does not have a bilingual version at all. In Table~\ref{tab:fail_rate}, we observe a high failure rate of around 60\% for almost all translated directions (except Zh$\rightarrow$En) due to translation and search failures, significantly exceeding the low failure rate of searching original entities online. Besides, even if we can find the right information of the translated entities, local people may not recognize or use the translated entities for communication, thus this results in inefficient communication with local people. 

\begin{table}[t!]
\setlength{\tabcolsep}{1.5pt}
\centering
\resizebox{\columnwidth}{!}{%
\begin{tabular}{cccccccc}
\toprule
Translate & Search     &En$\rightarrow$Zh & En$\rightarrow$Es & En$\rightarrow$Id   & Zh$\rightarrow$En & Es$\rightarrow$En & Id$\rightarrow$En \\
\midrule
\CheckmarkBold &  \CheckmarkBold & 35                  & 42                  & 36                  & 62                  & 30                  & 31                  \\
\CheckmarkBold &  \XSolidBrush   & 61                  & 34                  & 51                  & 18                  & 18                  & 15                  \\
 \XSolidBrush &  \CheckmarkBold   & 0                   & 24                  & 13                  & 11                  & 50                  & 54                  \\

\XSolidBrush   &  \XSolidBrush       & 4                   & 0                   & 0                   & 8                   & 2                   & 0                   \\
\midrule
\multicolumn{2}{c}{Failure Case (MTed Entities)}                    & 65                  & 58                  & 64                  & 37                  & 70                  & 69                  \\
\multicolumn{2}{c}{Failure Rate (MTed Entities)}          & 65\%                & 58\%                & 64\%                & 37\%                & 70\%                & 69\%                \\             
\multicolumn{2}{c}{Failure Rate (Original Entities)}                       & 3\%                & 3\%                & 3\%                & 0\%                & 1\%                & 0\%                \\
\bottomrule
\end{tabular}%
}
\caption{The search and translation results of 100 translated entities on Google. En$\rightarrow$Zh refers to the translation of English entities to Mandarin and Zh$\rightarrow$En refers to the translation of Mandarin entities to English.}
\label{tab:fail_rate}
\end{table}

\begin{figure}[t!]
\centering
\includegraphics[scale=0.5]{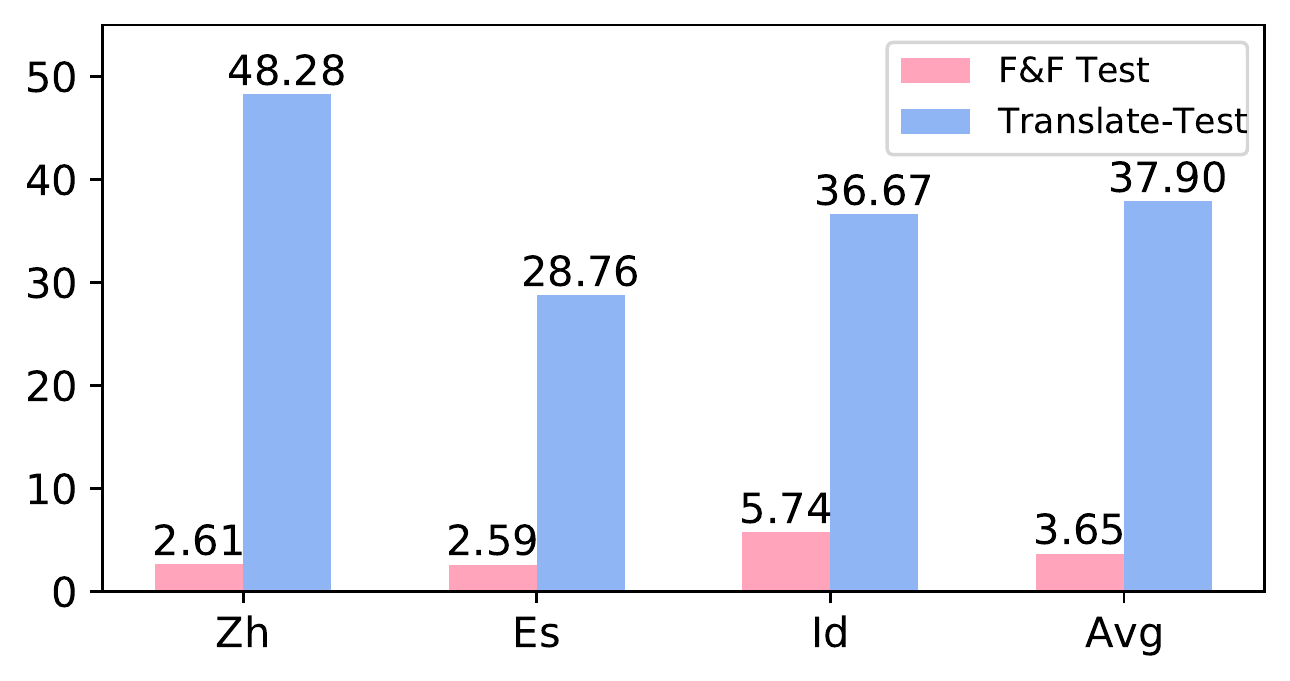}
\caption{ Joint accuracy of \notationb for DST on the \usecasea Test vs Translate-Test data.}
\label{fig:translate_test}
\end{figure}

\subsection{Overestimate of Translate-Train}
\label{sec:over-TT}
In previous translation-based work, a multilingual ToD system is usually built based on the translation of English training data (\notationb), and is evaluated on translated test data without any local entities (Translate-Test). To verify whether this procedure is reliable to build a multilingual ToD system, we also create a test dataset with translated entities instead of local entities in the target languages. As shown in Figure~\ref{fig:translate_test}, we find the \notationb model performs well on the test data with translated entities, but performs badly on the test data with real local entities. To the best of our knowledge, we provide the first analysis to identify this performance gap between the translated test data and data with real local entities in a more realistic use case \footnote{Please refer to Appendix~\ref{appendix:example_TT} for concrete examples where \notationb fails in predicting real local entities.}. Our work sheds light on the development of a globalized multilingual ToD system in practical use cases. We can tackle the challenge of localization issues by exploring new data augmentation method. Alternatively we can also explore new methods from the model level by building modular network to update the entities or perform transfer learning to adapt to new case without retraining.



\subsection{Local Context vs. Local Entities}
We compare the impact of training a model on data with either local contexts or local entities when the model is evaluated on monolingual test data in \usecasea and \usecased. Specifically, when the train set has access to local context only, all the entities in the train set are replaced by entities in non-target languages. Similarly, when the train set has access to local entities only, the contexts in the train set are replaced by context in the non-target languages. Table~\ref{tb:context_vs_entities} shows that both local contexts and local entities are essential to building ToD systems in the target language. A further analysis in Table~\ref{tb:context_vs_entities_language} and Table~\ref{tb:script_type} in the appendix shows that training with local entities is more important if the entities and contexts are written in the same type of language script (e.g. Latin script). 


\begin{table}[h!]
\centering
\resizebox{\columnwidth}{!}{%
\begin{tabular}{lccccc}
\toprule
\textbf{Train Set}  & \textbf{\usecased(en)} & \textbf{\usecasea(zh)} & \textbf{\usecasea(es)}  & \textbf{\usecasea(id)} &\textbf{avg}\\
\midrule
Local Context Only  & 5.46 & 1.77 & 2.37 & 2.40 & 3.20 \\
Local Entities Only & 6.39 & 0.36 & 2.41 & 2.75 & 3.05 \\
Local Context \& Entities & \textbf{52.78} & \textbf{36.97} & \textbf{24.66} &\textbf{25.26} & \textbf{38.13} \\
\bottomrule
\end{tabular}
}
\caption{Comparison of training with local context or/and local entities on the joint accuracy for DST in \usecased (en) and \usecasea (zh, es, id).}
\label{tb:context_vs_entities}
\end{table}

\subsection{Scaling up to 20 Languages}
With our proposed data curation method, it is possible to extend the dataset to cover more languages without spending extra costs if we skip the human post-editing step. Before doing so, one key question is whether the evaluation on the translated data without human post-editing is reliable as a proxy of the model performance. Thus, we conduct the experiments by evaluating the model performance of all baselines (\secref{sec:baselines}) on two sets of test data built with local entities: (1) \textbf{MT} test data where translated template is created by machine translation only~(\secref{sec:translate}); (2) \textbf{MTPE} test data where translated template is first translated by machines and post-edited later by professional translators. As shown in Table~\ref{tb:mt_test_result}, the overall reported results on MT test data are higher than those reported on MTPE test data, which is expected because the distribution of the MT test data is more similar to the MT training data. Although there are some differences on individual languages, the conclusions derived from the evaluations on the MT test data remain the same as those derived from the evaluation on the MTPE test data. We also calculate the Spearman rank correlation coefficient between the average results reported on MTPE test data and MT test data in Table~\ref{tb:mt_test_result}, which shows a statistically high correlation between the system performance on the MT test data and MTPE test data\footnote{Table~\ref{tb:spearman_breakdown} in the appendix shows detailed scores.}. Therefore, we show that the MT test data can be used as a proxy to estimate the model performance on the real test data for more languages. Thus we build MT test data for another 17 languages that are supported by Google Translate, Trip Advisor and Booking.com at the same time, as stated in Table~\ref{tb:selected_languages} and Table~\ref{tb:stats_ontology_17} in the appendix. Table~\ref{tb:global} shows the results of \notationa and \notationc on the test data of \usecasea, \usecaseb and \usecasec in 20 languages. The results show that the model has the best performance in the \usecaseb use case compared with the other two use cases, which is consistent with our findings in Table~\ref{tb:Use_case_result}.

\begin{table}[]
\centering
\resizebox{\columnwidth}{!}{%
\begin{tabular}{lcc|cc}
\toprule
\textbf{Use Case}  & \multicolumn{2}{c}{\textbf{F2F}}&\multicolumn{2}{|c}{\textbf{F2E}}\\
\midrule
\textbf{Methods}   & \textbf{MT Test} & \textbf{MTPE Test} & \textbf{MT Test} & \textbf{MTPE Test} \\
\midrule
\notationa         &  1.29 & 1.28  &  9.64  & 9.12  \\
\notationb         &  3.71 & 3.65  &  4.17 & 3.97  \\
\notationc         &  35.78 & 28.96 &  56.15 & 48.71 \\
\notationd         &  36.31 & 29.74 &  57.84 & 54.29 \\
\notatione         &  \textbf{37.89} & \textbf{30.76} &  \textbf{58.76} & \textbf{56.28} \\
\midrule
Spearman's correlation & \multicolumn{2}{c}{\textbf{1.0}}&\multicolumn{2}{|c}{\textbf{1.0}}\\
\bottomrule
\end{tabular}
}
\caption{Comparison of average joint accuracy on DST reported on MT test data and MTPE test data for use case \usecasea and \usecaseb}
\label{tb:mt_test_result}
\end{table}

\begin{table}[t!]
\centering
\scalebox{0.7}{
\begin{tabular}{lcc}
\toprule
 \textbf{Case} &\textbf{Method} &  \textbf{Avg} \\
 
\midrule
\multirow{2}{*}{\usecasea} &\notationa  & 1.48 \\

& \notationc & \textbf{16.12} \\

\midrule
\multirow{2}{*}{\usecaseb} &\notationa  &  9.03 \\

& \notationc & \textbf{34.20} \\
\midrule
\multirow{2}{*}{\usecasec} &\notationa  & 1.97 \\

& \notationc & \textbf{23.40}  \\ 

\bottomrule
\end{tabular}}
\caption{Average results of \notationa on test data of \usecasea, \usecaseb and \usecasec in 20 languages. Please refer to Table~\ref{tb:global_breakdown} in Section~\ref{appendix:20languages} for the break down results of 20 languages. }
\label{tb:global}
\end{table}

%% file: 02_related_work.tex
\section{Related Work}
\label{sec:related_work}


 Over the last few years, the success of ToD systems is largely driven by the joint advent of neural network models \citep{eric-etal-2017-key,wu-etal-2019-transferable,lin-etal-2020-mintl} and collections of large-scale annotation corpora. These corpora cover a wide range of topics from a single domain (e.g., ATIS \cite{hemphill1990atis}, DSTC 2 \cite{henderson2014second}, Frames \cite{el-asri-etal-2017-frames}, KVRET \cite{eric-etal-2017-key}, WoZ 2.0 \cite{wen-etal-2017-network}, M2M \cite{schatzmann2007agenda}) to multiple domains (e.g., MultiWoZ \cite{budzianowski-etal-2018-multiwoz}, SGD \cite{rastogi2020towards}). Most notably among these collections, MultiWoZ is a large-scale multi-domain dataset that focuses on transitions between different domains or scenarios in real conversations \citep{budzianowski-etal-2018-multiwoz}. Due to the high cost of collecting task-oriented dialogues, only a few monolingual or bilingual non-English ToD datasets are available \cite{zhu2020crosswoz, quan-etal-2020-risawoz, lin2021bitod}. While there is an increasing interest in data curation for multilingual ToD systems, a vast majority of existing multilingual ToD datasets do not consider the real use cases when using a ToD system to search for local entities in a country. We fill this gap in this paper to provide the first analysis on three previously unexplored use cases.

%% file: 08_conclusion.tex
\section{Conclusions}
\label{sec:conclusion}
In this paper, we provide an analysis on three unexplored use cases for multilingual task-oriented dialogue systems. We propose a new data curation method that leverages a machine translation system and local entities in target languages to create a new multilingual TOD dataset,~\name. We propose a series of strong baseline methods and conduct extensive experiments on \name to encourage research for multilingual ToD systems. Besides, we extend the coverage of languages on multilingual ToD to 20 languages, marking the one step further towards building a globalized multilingual ToD system for all of the world’s citizen. 

\section{Ethical Review}
 In this section, we would like to address the ethical concerns. All the professional translators in this project have been properly compensated. For Chinese and Spanish, we have followed the standard procurement requirements and engaged three translation companies for quality and price comparison. A small sample of the data had been given to them for MTPE and we then compared their translation results. Following that, we selected the company that produced the best sample translation, and submitted the full translation orders according to the agreed price quotations. For Indonesian, three translation companies were also requested to provide sample MTPE, but our quality check found the quality of these samples to be unsatisfactory. So, no company was engaged, and our in-house Indonesian linguistic resources were used instead. These Indonesian linguists were assigned to work on this project during normal working hours and given proper compensation complying with the local labor laws. 


\section*{Acknowledgements}
\label{sec:acknowledgements}
This research is partly supported by the Alibaba-NTU Singapore Joint Research Institute, Nanyang Technological University. All the costs for machine translation post-editing are funded by DAMO Academy, Alibaba Group. We would like to thank the help from our Alibaba colleagues, Haiyun Peng, Zifan Xu and Ruidan He, and our NTU-NLP team member, Chengwei Qin in this work as well.

%% file: 09_appendix.tex
\section*{Appendix}

\section{Comparison of Four Use Cases}\label{appendix:comparison}

\begin{table}[h!]
\centering
\scalebox{0.8}{
\begin{tabular}{cccc}
\toprule
\multirow{2}{*}{Use Case} & \multirow{2}{*}{Source ToD} & Speaker & Country \\
    &         & (ToD Context) & (ToD Ontology) \\ \midrule
\usecasea & \multirow{4}{*}{English} & Foreign Lang. & Foreign Lang. \\ 
\usecaseb &  & Foregin Lang. & English       \\ 
\usecasec &  & English       & Foreign Lang. \\ 
\usecased &  & English       & English       \\
\bottomrule
\end{tabular}
}
\caption{Four use cases of multilingual ToD systems: A foreign language or English speaker travels to a country of a foreign language or English. }
\label{tb:use-case}
\end{table}

\section{Examples of Labeled Sequence Translation}\label{appendix:sequence_translation}

\begin{figure*}[h!]
\centering
\includegraphics[scale=0.2]{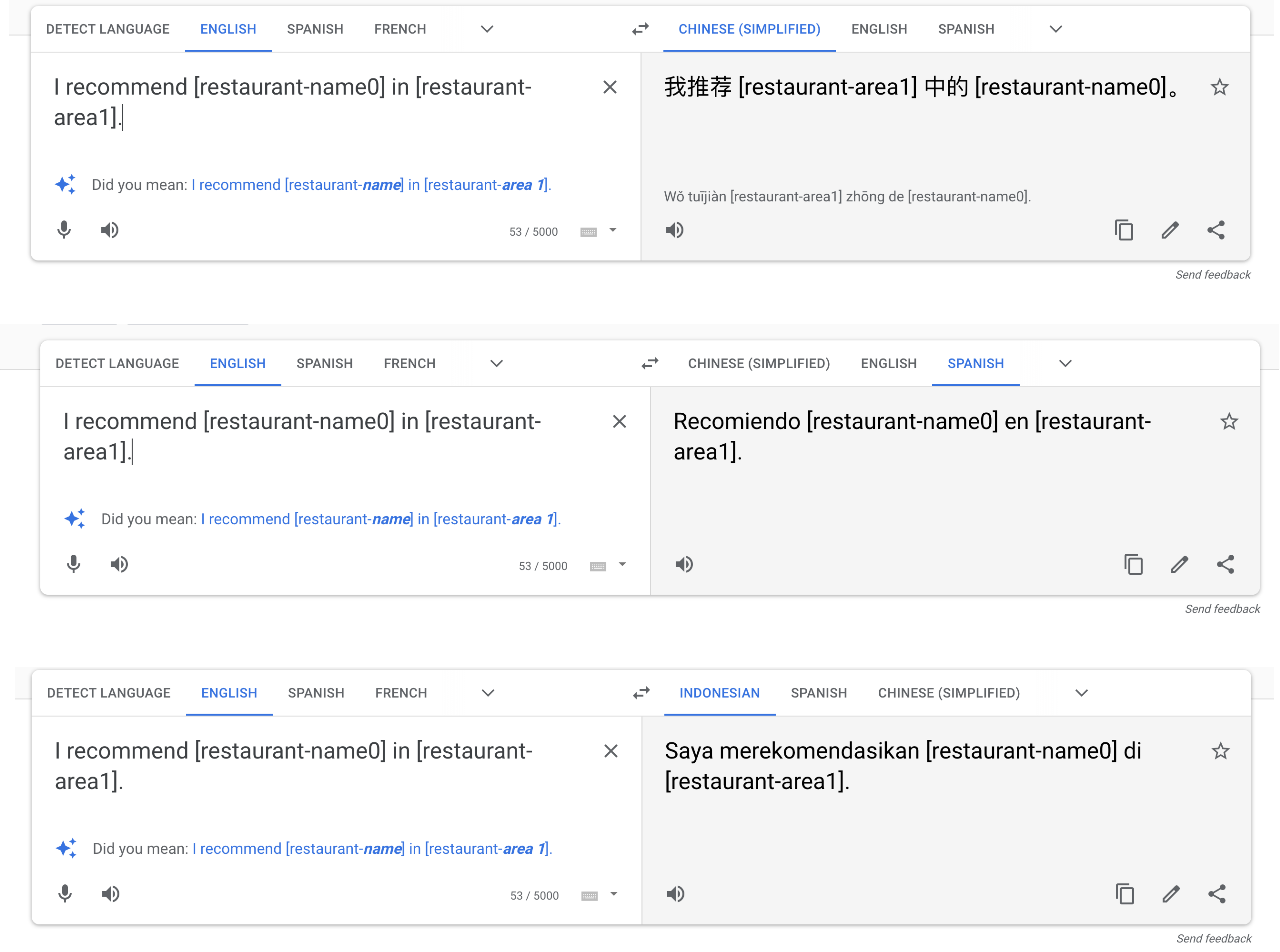}
\caption{An instance of labeled sequence translation with google translate, from English to three target languages, Mandarin, Spanish and Indonesian.}
\label{fig:sequence_translation}
\end{figure*}

\section{BLEU Score of MT versus MTPE Test Template} \label{appendix:bleu}
\begin{table}[h!]
\centering
\scalebox{0.8}{
\begin{tabular}{lcccc}
\toprule
Languages & Zh & Es    & Id & Avg \\
\midrule
BLEU Score & 55.61 &	49.33 &	48.97 &	51.30    \\
 \bottomrule
\end{tabular}}
\caption{BLEU Scores of MT Test Template using MTPE Test Template as reference.}
\end{table}

\newpage
\section{Test Set Distribution} \label{appendix:test_dist}
\begin{figure*}[h!]
\centering
\includegraphics[scale=0.5]{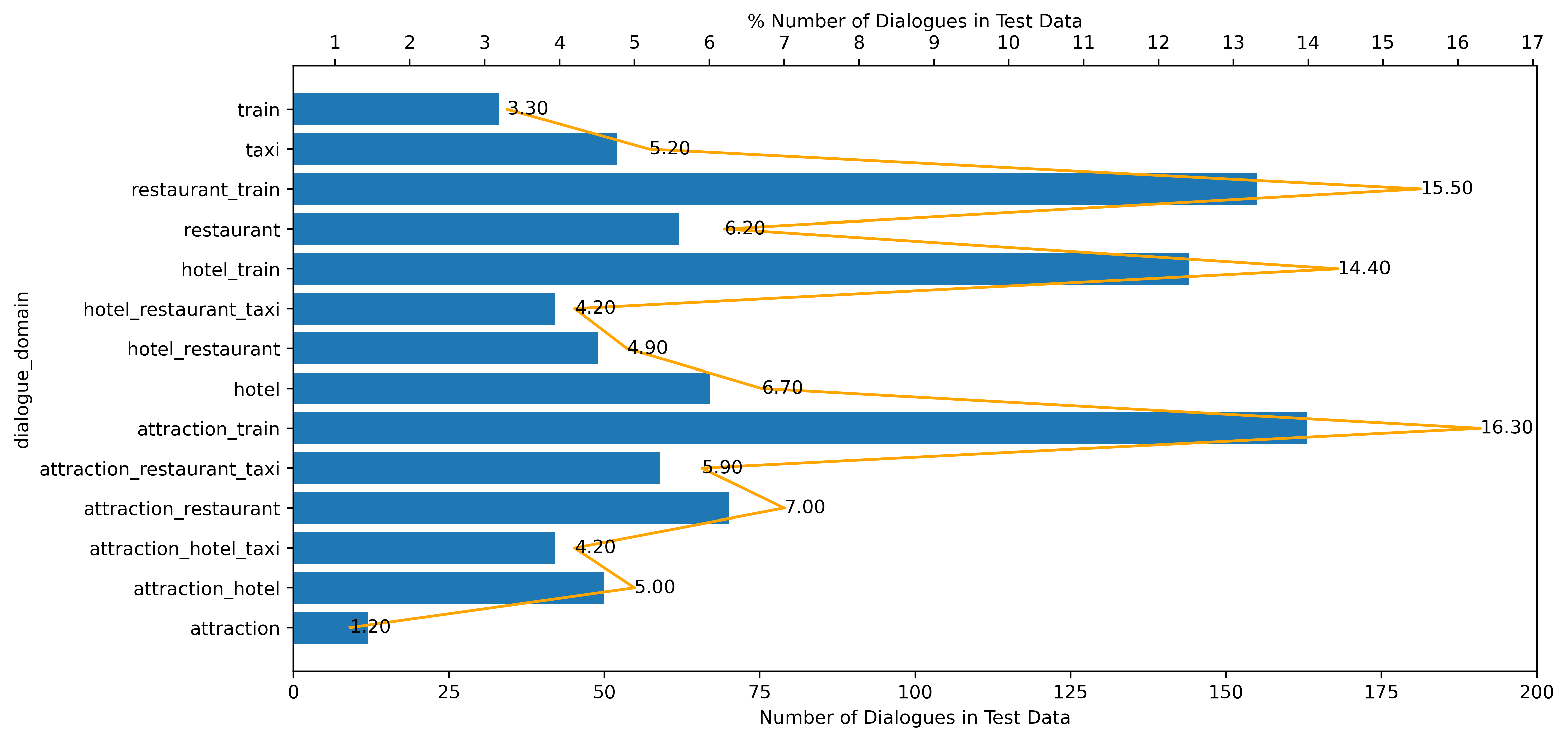}
\caption{Gold English Test Set Distribution by Domains. We follow this distribution to select the top 500 high-scoring dialogues in the test set for post-editing.}
\label{fig:test_distribution}
\end{figure*}

\section{Selected Languages} \label{appendix:sel_lang}
\begin{table*}[h!]
\resizebox{\columnwidth}{!}{%
\centering
\begin{tabular}{lcccccc}
\toprule
Language   & ISO639-1code & Language Family & \# Wikipedia articles (in millions) & High / Middle/ Low  Resource & Writing Script & Selected City  \\
\midrule
English    & en & IE: Germanic   & 6.35 & High   & Latin             & Cambridge        \\
Swedish    & sv & IE: Germanic   & 2.95 & High   & Latin             & Stockholm        \\
German     & de & IE: Germanic   & 2.61 & High   & Latin             & Berlin           \\
French     & fr & IE: Romance    & 2.35 & High   & Latin             & Paris            \\
Dutch      & nl & IE: Germanic   & 2.06 & High   & Latin             & Amsterdam        \\
Russian    & ru & IE: Slavic     & 1.74 & High   & Cyrillic          & Moscow           \\
Italian    & it & IE: Romance    & 1.71 & High   & Latin             & Rome             \\
Spanish    & es & IE: Romance    & 1.71 & High   & Latin             & Barcelona        \\
Japanese   & ja & Japonic        & 1.28 & High   & Ideograms         & Tokyo            \\
Vietnamese & vi & Austro-Asiatic & 1.27 & High   & Latin             & Ho Chi Minh City \\
Mandarin    & zh & Sino-Tibetan   & 1.22 & High   & Chinese ideograms & Shanghai         \\
Arabic     & ar & Afro-Asiatic   & 1.13 & High   & Arabic            & Cairo            \\
Portuguese & pt & IE: Romance    & 1.07 & High   & Latin             & Lisbon           \\
Indonesian & id & Austronesian   & 0.59 & Middle & Latin             & Jakarta          \\
Norwegian  & no & IE: Germanic   & 0.56 & Middle & Latin             & Oslo             \\
Korean     & ko & Koreanic       & 0.55 & Middle & Hangul            & Seoul            \\
Turkish    & tr & Turkic         & 0.42 & Middle & Latin             & İstanbul         \\
Hebrew     & he & Afro-Asiatic   & 0.30 & Low    & Hebrew            & Tel Aviv         \\
Danish     & da & IE: Germanic   & 0.27 & Low    & Latin             & Copenhagen       \\
Greek      & el & IE: Greek      & 0.20 & Low    & Greek             & Athens           \\
Thai       & th & Kra-Dai        & 0.14 & Low    & Brahmic           & Bangkok       \\
\bottomrule
\end{tabular}
}
\caption{Statistics about languages in the cross-lingual benchmark. The selected 21 languages (including English) belong to 8 language families and 1 isolate, with
Indo-European (IE) having the most members. We categorize the languages with more than 1 million, more than 400 thousand but less than 1 million, less than 400 thousand Wikipedia articles as high resource languages, middle resource languages and low resource languages. For each language, we select one city for each language to collect localized ontology. }
\label{tb:selected_languages}
\end{table*}

\newpage
\section{Statistics of Entities in the Collected Ontology} \label{appendix:stat_ontology}
\begin{table}[h!]
\centering
\scalebox{0.78}{
\begin{tabular}{lccccccc}
\toprule
\textbf{Languages}          & \textbf{rest.} & \textbf{hotel} & \textbf{attr.} & \textbf{train}  &  \textbf{taxi}  \\
\midrule
en        & 110        & 33    & 79         & 2828  & 222  \\
zh        & 3000       & 496   & 1000       & 100   & 4496 \\
es        & 3000       & 426   & 1000       & 100   & 4426 \\
id        & 3000       & 999   & 792        & 100   & 4791 \\
ar        & 2989       & 680   & 1000       & 100   & 4669 \\
da        & 2343       & 165   & 1000       & 100   & 3508 \\
de        & 2988       & 659   & 1000       & 100   & 4647 \\
el        & 2600       & 1000  & 1000       & 100   & 4600 \\
fr        & 3000       & 1000  & 1000       & 100   & 5000 \\
he        & 1558       & 258   & 1000       & 100   & 2258 \\
it        & 3000       & 800   & 1000       & 100   & 2800 \\
ja        & 2967       & 864   & 1000       & 100   & 4831 \\
ko        & 2990       & 532   & 1000       & 100   & 4522 \\
nl        & 2990       & 537   & 1000       & 100   & 4527 \\
no        & 1293       & 95    & 757        & 100   & 2145 \\
pt        & 2993       & 951   & 1000       & 100   & 4944 \\
ru        & 2985       & 531   & 1000       & 100   & 4516 \\
sv        & 3000       & 214   & 891        & 100   & 4105 \\
th        & 2995       & 1000  & 1000       & 100   & 4995 \\
tr        & 2986       & 533   & 1000       & 100   & 4519 \\
vi        & 2991       & 773   & 1000       & 100   & 4764\\
\bottomrule
\end{tabular}
}
\caption{Statistics of entities in the collected ontology in different languages. We count the number of entities in the database of each domain. Noticed that in the Taxi database of MultiWoZ, it only list down the taxi colors, taxi types and taxi phones. The taxi destination and departure refer to the entities in the restaurant, hotel and attraction domains. Thus, we use the sum of the number of entities in Restaurant, Hotel and Attraction domains as a proxy of the total number of entities in taxi domain. Besides, we follow MultiWoZ to collect one hospital and one police station for each city.}
\label{tb:stats_ontology_17}
\end{table}

\section{Statistics of \name} \label{appendix:data_stat}
\begin{table}[h!]
\centering
\resizebox{\columnwidth}{!}{%
\begin{tabular}{lcccccccccccc}
\toprule
Use Case  & \multicolumn{4}{l}{F\&F}              & \multicolumn{4}{l}{F\&E}              & \multicolumn{4}{l}{E\&F}              \\
Languages & Train \& Dev & Method & Test & Method & Train \& Dev & Method & Test & Method & Train \& Dev & Method & Test & Method \\
\midrule
zh                   & 9438                 & MT                   & 1000                 & MTPE                 & 9438                 & MT                   & 1000                 & MTPE                 & 9438                 & Human                & 1000                 & Human                \\
es                   & 9438                 & MT                   & 1000                 & MTPE                 & 9438                 & MT                   & 1000                 & MTPE                 & 9438                 & Human                & 1000                 & Human                \\
id                   & 9438                 & MT                   & 1000                 & MTPE                 & 9438                 & MT                   & 1000                 & MTPE                 & 9438                 & Human                & 1000                 & Human                \\
ar                   & 9438                 & MT                   & 1000                 & MT                   & 9438                 & MT                   & 1000                 & MT                   & 9438                 & Human                & 1000                 & Human                \\
da                   & 9438                 & MT                   & 1000                 & MT                   & 9438                 & MT                   & 1000                 & MT                   & 9438                 & Human                & 1000                 & Human                \\
de                   & 9438                 & MT                   & 1000                 & MT                   & 9438                 & MT                   & 1000                 & MT                   & 9438                 & Human                & 1000                 & Human                \\
el                   & 9438                 & MT                   & 1000                 & MT                   & 9438                 & MT                   & 1000                 & MT                   & 9438                 & Human                & 1000                 & Human                \\
fr                   & 9438                 & MT                   & 1000                 & MT                   & 9438                 & MT                   & 1000                 & MT                   & 9438                 & Human                & 1000                 & Human                \\
he                   & 9438                 & MT                   & 1000                 & MT                   & 9438                 & MT                   & 1000                 & MT                   & 9438                 & Human                & 1000                 & Human                \\
it                   & 9438                 & MT                   & 1000                 & MT                   & 9438                 & MT                   & 1000                 & MT                   & 9438                 & Human                & 1000                 & Human                \\
ja                   & 9438                 & MT                   & 1000                 & MT                   & 9438                 & MT                   & 1000                 & MT                   & 9438                 & Human                & 1000                 & Human                \\
ko                   & 9438                 & MT                   & 1000                 & MT                   & 9438                 & MT                   & 1000                 & MT                   & 9438                 & Human                & 1000                 & Human                \\
nl                   & 9438                 & MT                   & 1000                 & MT                   & 9438                 & MT                   & 1000                 & MT                   & 9438                 & Human                & 1000                 & Human                \\
no                   & 9438                 & MT                   & 1000                 & MT                   & 9438                 & MT                   & 1000                 & MT                   & 9438                 & Human                & 1000                 & Human                \\
pt                   & 9438                 & MT                   & 1000                 & MT                   & 9438                 & MT                   & 1000                 & MT                   & 9438                 & Human                & 1000                 & Human                \\
ru                   & 9438                 & MT                   & 1000                 & MT                   & 9438                 & MT                   & 1000                 & MT                   & 9438                 & Human                & 1000                 & Human                \\
sv                   & 9438                 & MT                   & 1000                 & MT                   & 9438                 & MT                   & 1000                 & MT                   & 9438                 & Human                & 1000                 & Human                \\
th                   & 9438                 & MT                   & 1000                 & MT                   & 9438                 & MT                   & 1000                 & MT                   & 9438                 & Human                & 1000                 & Human                \\
tr                   & 9438                 & MT                   & 1000                 & MT                   & 9438                 & MT                   & 1000                 & MT                   & 9438                 & Human                & 1000                 & Human                \\
vi                   & 9438                 & MT                   & 1000                 & MT                   & 9438                 & MT                   & 1000                 & MT                   & 9438                 & Human                & 1000                 & Human           \\
\bottomrule
\end{tabular}
}
\caption{Statistics of created dataset, \name for each use case in each target language. For \usecasec, as the context is the original Engish data, we consider it is created by human. For test data of zh, es and id, we replace the entities twice to boostrap the test data to 1000 dialogues. We are currently preparing the post editing of the other 500 dialogues in test data. Meanwhile, we are leveraging machine translation to prepare the train data for the 17 languages and will release it with baselines in the next version soon.}
\label{tb:stats_globalwoz}
\end{table}

\section{Dialogue Examples} \label{appendix:dialogue_example}
\begin{figure*}[h!]
\centering
\includegraphics[width=\columnwidth]{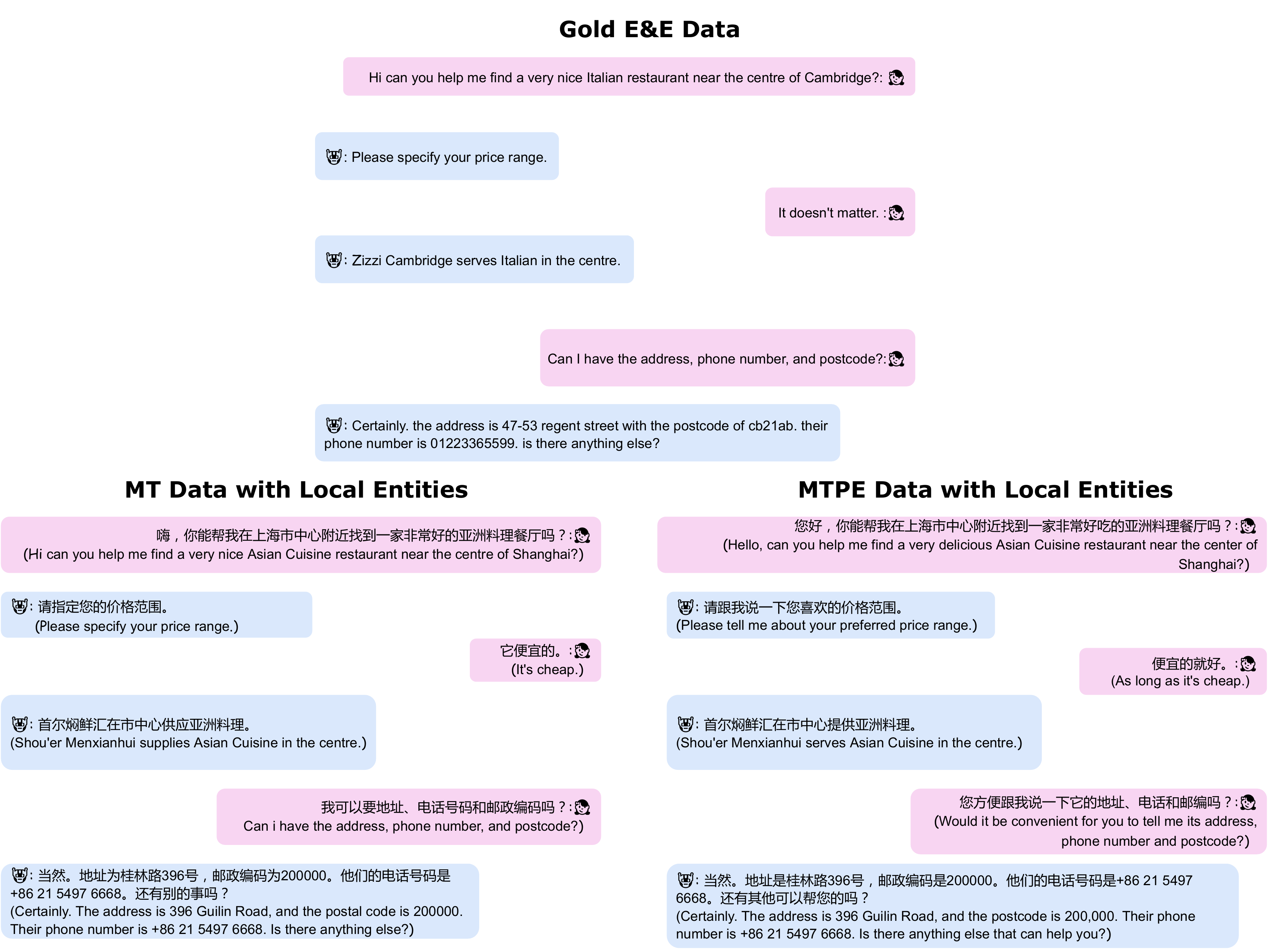}
\caption{Examples of some utterances in original \usecased data, MT data and MTPE data,}
\label{fig:example}
\end{figure*}

\newpage
\section{Summary of Proposed Baselines}\label{appendix:summary_baselines}
\begin{table}[h!]
\centering
\scalebox{0.7}{
\begin{tabular}{lccccc}
\toprule
\textbf{Methods}          &  En Context            & En Entities           & Local Context         & Local Entities        & Translated Entites    \\
\midrule
\notationa  &   \CheckmarkBold & \CheckmarkBold &  & &   \\
\notationb  &    &    & \CheckmarkBold & & \CheckmarkBold \\
\midrule
\notationc (\usecasea)     &    &    & \CheckmarkBold &  \CheckmarkBold & \\
\notationc (\usecaseb)     &    &  \CheckmarkBold   & \CheckmarkBold &  & \\
\notationc (\usecasec)     &  \CheckmarkBold   &    &  & \CheckmarkBold & \\
\bottomrule
\end{tabular}
}
\caption{Accessibility of different types of context and entities for each method.}
\label{tb:Methods_data}
\end{table}

\begin{table}[h!]
\centering
\scalebox{0.8}{
\begin{tabular}{lcccc}
\toprule
\textbf{Methods}          &\usecased   &  \usecasea            & \usecaseb      & \usecasec         \\
\midrule
\notationa  & \CheckmarkBold    & &  &  \\
\notationb  &    &    &  &   \\
\midrule
\notationc (\usecasea)     &   &  \CheckmarkBold   &    &   \\
\notationc (\usecaseb)     &    &    &  \CheckmarkBold &   \\
\notationc (\usecasec)     &   &    &   &  \CheckmarkBold  \\
\midrule
\notationd (\usecased+ \usecasea)     & \CheckmarkBold   &  \CheckmarkBold   &    &   \\
\notationd (\usecased+ \usecaseb)     &  \CheckmarkBold   &    &  \CheckmarkBold &   \\
\notationd (\usecased+ \usecasec)    & \CheckmarkBold   &    &   &  \CheckmarkBold  \\
\midrule
\notatione (\usecased+ \usecasea) &  \CheckmarkBold   &  \CheckmarkBold   &    &    \\
\notatione (\usecased+ \usecaseb) &  \CheckmarkBold   &    &  \CheckmarkBold & \\
\notatione (\usecased+ \usecasec) &  \CheckmarkBold   &    &  & \CheckmarkBold  \\
\midrule
\notationf (\usecased+ \usecasea + \usecaseb + \usecasec) &  \CheckmarkBold   & \CheckmarkBold    & \CheckmarkBold  & \CheckmarkBold \\
\bottomrule
\end{tabular}
}
\caption{Accessibility of data in each use case for each method. Noticed that \notationb doesn't have access to the data of the four use cases. \notationb has access to a set of pseudo-labeled training data created by replacing the placeholders in the translated template with machine-translated entities instead of local entities.}
\label{tb:Methods_data_access}
\end{table}

\section{Use Case \usecased} \label{appendix:EE}
We also compare the performance of all methods on the original \usecased test data. As \textbf{\notationa} is trained on monolingual English training data, it gets a high accuracy of 52.78 on the English test data. In contrast, \textbf{\notationb} and \textbf{\notationc} (\usecasea) perform poorly on the English test data, because both of them have no access to any English data. Comparing to \notationC (\usecasea), \textbf{\notationc} (\usecaseb) and \textbf{\notationc} (\usecasec) achieve higher accuracy scores as they either have access to English context or English entities. When we perform bilingual and multilingual joint training (i.e., \textbf{\notationd} and \textbf{\notatione}), the base model has a performance increase except \textbf{\notatione} (\usecased+ \usecasec). This shows that bilingual and multilingual joint training may be used to improve the performance on source language. Further research can be done in this line. 

\begin{table}[h!]
\centering
\scalebox{0.7}{
\begin{tabular}{lc}
\toprule
\textbf{Methods}          & \textbf{En}\\
\midrule
\notationa  &   52.78  \\
 \notationb          &  2.27  \\
\midrule
\notationc (\usecasea)      &  1.09 \\
\notationc (\usecaseb)      & 6.39 \\
\notationc (\usecasec)      &  5.46 \\
\midrule
\notationd (\usecased+ \usecasea) &     52.87  \\
\notationd (\usecased+ \usecaseb) &     \textbf{53.69}     \\
\notationd (\usecased+ \usecasec) &     53.05    \\
\midrule
\notatione (\usecased+ \usecasea) &  53.28  \\
\notatione (\usecased+ \usecaseb) &  53.43 \\
\notatione (\usecased+ \usecasec) &  51.75  \\
\bottomrule
\end{tabular}
}
\caption{Joint accuracy on DST in three target languages on the English test data. 
}
\label{tb:En_result}
\end{table}

\newpage
\section{Breakdown of Few Shot Results}
\begin{table}[h!]
\centering
\scalebox{0.7}{
\begin{tabular}{lcccc}
\toprule
\multicolumn{5}{c}{Zero Shot (E\&E)}   \\
\midrule
Use Case     & Zh       & Es      & Id      & Avg     \\
\midrule
F2F          & 1.22     & 1.38    & 1.26    & 1.28    \\
F2E          & 6.92     & 11.34   & 9.09    & 9.12    \\
E2F          & 1.69     & 1.81    & 1.82    & 1.77    \\
\midrule
\multicolumn{5}{c}{Few Shot + Zero Shot (E\&E)}       \\
\midrule
Use Case     & Zh       & Es      & Id      & Avg     \\
\midrule
F2F          & 15.93    & 7.13    & 12.09   & 11.72   \\
F2E          & 39.88    & 39.38   & 43.26   & 40.84   \\
E2F          & 20.61    & 14.17   & 18.55   & 17.78   \\
\midrule
\multicolumn{5}{c}{SUC}                               \\
\midrule
Use Case     & Zh       & Es      & Id      & Avg     \\
\midrule
F2F          & 36.97    & 24.66   & 25.26   & 28.96   \\
F2E          & 56.28    & 41.94   & 47.93   & 48.71   \\
E2F          & 38.56    & 28.00   & 43.82   & 36.79   \\
\midrule
\multicolumn{5}{c}{Few Shot + SUC}                    \\
\midrule
Use Case     & Zh       & Es      & Id      & Avg     \\
\midrule
F2F          & 37.81    & 25.15   & 39.51   & 34.16   \\
F2E          & 58.39    & 53.03   & 54.02   & 55.15   \\
E2F          & 38.75    & 27.66   & 44.23   & 36.88   \\
\midrule
\multicolumn{5}{c}{Few Shot + Zero Shot (E\&E) + SUC} \\
\midrule
Use Case     & Zh       & Es      & Id      & Avg     \\
\midrule
F2F          & 37.52    & 26.44   & 40.15   & 34.70   \\
F2E          & 59.21    & 54.93   & 56.17   & 56.77   \\
E2F          & 39.51    & 27.84   & 45.48   & 37.61  \\
\bottomrule
\end{tabular}
}
\caption{A breakdown of few-shot cross-lingual average joint accuracy on DST over three target languages in three use cases.}
\label{tb:few_shot_breakdown}
\end{table}
\newpage

\section{Concrete Examples where Translate-Train Performs Badly on the Test Data with Real Local Entities.} \label{appendix:example_TT}

Through investigation, we found that the \notationb method usually performed badly in two main scenarios. Figure~\ref{fig:example} is the illustrations of the two scenarios. Scenario 1 is when the \notationb can predict values that are close to the meaning of the ground truth values but suffer from the problems of translationese. For example, model trained with \notationb may predict "\cchar{美食酒吧}" (gastropub), which is a direct translation of gastropub and not commonly used in Chinese instead of "\cchar{酒吧餐}" (bar).  Scenario 2 is when \notationb needs to predict the name of real localized entities which \notationb doesn't have access to. For example, trained with \notationb may predict "\cchar{冈维尔酒店}" (Gonville Hotel) which is a direct translation of Gonville Hotel, instead of "\cchar{汉庭酒店}" (Hanting Hotel) which is unseen in \notationb training data.

\begin{figure*}[h!]
\centering
\includegraphics[scale=0.9]{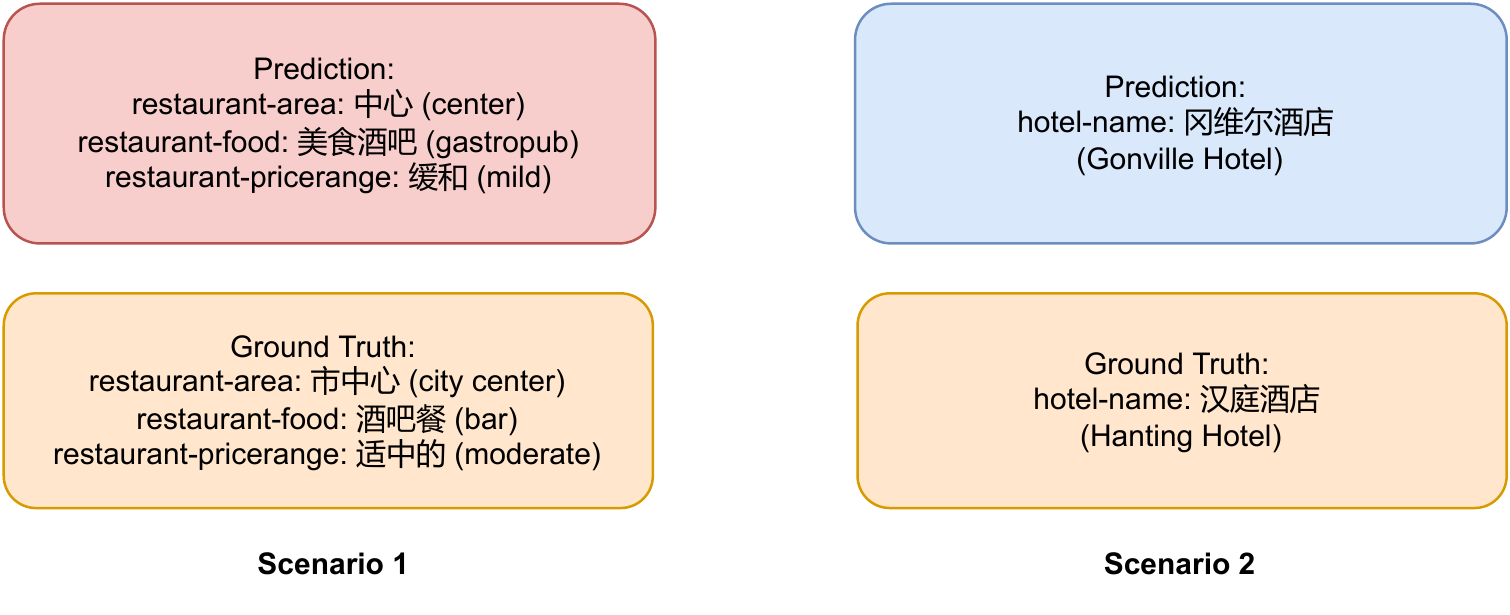}
\caption{Concrete examples where Translate-Train performs badly on the test data with real local entities. 
}
\label{fig:example}
\end{figure*}

\section{Breakdown of the Results of Local Context vs Local Entities by Languages}
\begin{table}[h!]
\centering
\scalebox{0.75}{
\begin{tabular}{lccccc}
\toprule
\multicolumn{5}{c}{\usecased(en)}                          \\
\midrule
Context vs Entities & \cellcolor{lavender} Zh   & Es   & Id   & Avg  \\
En\_Context         & \cellcolor{lavender} 5.37 & 5.33 & 5.67 & 5.46 \\
En\_Entites         & \cellcolor{lavender} 3.49 & 7.78 & 7.90 & 6.39 \\
\midrule
\multicolumn{5}{c}{\usecasea(zh)}                           \\
\midrule
Context vs Entities & \cellcolor{lavender} En   & \cellcolor{lavender} Es   & \cellcolor{lavender} Id   & Avg  \\
Zh\_Context         &\cellcolor{lavender}  1.74 & \cellcolor{lavender} 1.77 &\cellcolor{lavender}  1.80 & 1.77 \\
Zh\_Entites         &\cellcolor{lavender}  0.27 &\cellcolor{lavender}  0.73 & \cellcolor{lavender} 0.10 & 0.36 \\
\midrule
\multicolumn{5}{c}{\usecasea(es)}                          \\
\midrule
Context vs Entities & En   & \cellcolor{lavender} Zh   & Id   & Avg  \\
Es\_Context         & 1.73 & \cellcolor{lavender} 2.01 & 3.37 & 2.37 \\
Es\_Entites         & 3.92 &\cellcolor{lavender}  0.44 & 2.86 & 2.41 \\
\midrule
\multicolumn{5}{c}{\usecasea(id)}                           \\
\midrule
Context vs Entities & En   & \cellcolor{lavender} Zh   & Es   & Avg  \\
Id\_Context         & 2.07 &\cellcolor{lavender}  2.18 & 2.94 & 2.40 \\
Id\_Entites         & 3.92 & \cellcolor{lavender} 0.84 & 3.48 & 2.75 \\
\bottomrule
\end{tabular}
}
\caption{A breakdown of comparison of the impact of local context and local entities on joint accuracy for DST in each language. The cases where context and entities are in different script types are highlighted in lavender color.}
\label{tb:context_vs_entities_language}
\end{table}

\begin{table}[h!]
\centering
\scalebox{0.7}{
\begin{tabular}{lcc}
\toprule
\textbf{Train Set}  & \textbf{different script type} & \textbf{same script type} \\
\midrule
Local Context Only  & \textbf{2.48}             & 3.52                  \\
Local Entities Only & 0.98             & \textbf{4.98}                  \\ 
\bottomrule
\end{tabular}
}
\caption{Comparison of the impact of script type on Local Context Only vs Local Entities Only. It shows that training with local entities is more important if the entities and contexts are written in the same type of language script (e.g. Latin script), otherwise training with local contexts is more important.}
\label{tb:script_type}
\end{table}

\newpage
\section{Breakdown of MT Test Data vs MTPE Test Data by Languages}
\begin{table}[h!]
\centering
\scalebox{0.75}{
\begin{tabular}{lcccccc}
\toprule
Languages          & \multicolumn{2}{c}{Zh}   & \multicolumn{2}{c}{Es}   & \multicolumn{2}{c}{Id}   \\
\midrule
F2F                & MT          & MTPE       & MT          & MTPE       & MT          & MTPE       \\
\midrule
\notationa     & 1.19        & 1.22       & 1.40        & 1.38       & 1.28        & 1.26       \\
\notationb     & 2.50        & 2.61       & 2.81        & 2.59       & 5.81        & 5.74       \\
\notationc     & 37.79       & 36.97      & 26.95       & 24.66      & 42.59       & 25.26      \\
\notationd     & 38.62       & 37.32      & 27.34       & 25.52      & 42.96       & 26.39      \\
\notatione     & 39.11       & 38.01      & 29.17       & 26.03      & 45.39       & 28.22   \\
\midrule
Spearman’s correlation          & \multicolumn{2}{c}{1.00} & \multicolumn{2}{c}{1.00} & \multicolumn{2}{c}{1.00} \\
\midrule
F2E                & MT          & MTPE       & MT          & MTPE       & MT          & MTPE       \\
\midrule
\notationa         & 7.61        & 6.92       & 11.67       & 11.34      & 9.64        & 9.09       \\
\notationb         & 2.25        & 2.28       & 5.25        & 4.97       & 5.03        & 4.67       \\
\notationc         & 57.10       & 56.28      & 55.70       & 41.94      & 55.64       & 47.93      \\
\notationd         & 59.05       & 59.87      & 57.68       & 48.20      & 56.80       & 54.79      \\
\notatione         & 60.48       & 60.37      & 57.04       & 53.56      & 58.23       & 54.93 \\
\midrule
Spearman’s correlation          & \multicolumn{2}{c}{1.00} & \multicolumn{2}{c}{0.90} & \multicolumn{2}{c}{1.00}     \\
\bottomrule
\end{tabular}
}
\caption{Spearman rank correlation coefficient between the results on MTPE test data and MT test data for each language.}
\label{tb:spearman_breakdown}
\end{table}


\section{Breakdown of Results of 20 Languages} \label{appendix:20languages}
\begin{table*}[h!]
\centering
\scalebox{0.5}{
\begin{tabular}{lcccccccccccccccccccccc}
\toprule
 \textbf{Case} &\textbf{Method} & \cellcolor{palepink} \textbf{zh} & \cellcolor{palepink} \textbf{es} & \cellcolor{palepink} \textbf{id} & \textbf{ar} & \textbf{da} & \textbf{de} & \textbf{el}  & \textbf{fr} & \textbf{he} & \textbf{it} & \textbf{ja} & \textbf{ko} & \textbf{nl} & \textbf{no} & \textbf{pt} & \textbf{ru} & \textbf{sv} & \textbf{th} & \textbf{tr} & \textbf{vi}  & \textbf{avg} \\
 
\midrule
\multirow{2}{*}{\usecasea} &\notationa  & \cellcolor{palepink} 1.22 & \cellcolor{palepink} 1.38 & \cellcolor{palepink} 1.26 & 1.49 & 1.52 & 1.52 & 1.51 & 2.04 & 1.47 & 1.55 & 1.48 & 1.51 & 1.55 & 1.51 & 1.53 & 1.52 & 1.41 & 1.57 & 1.22 & 1.41 & 1.48 \\

& \notationc &  \cellcolor{palepink}\textbf{36.97} & \cellcolor{palepink}\textbf{24.66} &\cellcolor{palepink} \textbf{25.26} & \textbf{14.33} & \textbf{24.08} & \textbf{15.31} & \textbf{4.33}  & \textbf{23.72} & \textbf{7.76}  & \textbf{18.81} & \textbf{20.98} & \textbf{1.71}  & \textbf{23.87} & \textbf{24.86} & \textbf{14.91} & \textbf{13.00} & \textbf{11.31} & \textbf{2.74}  & \textbf{10.65} & \textbf{3.06}  & \textbf{16.12} \\

\midrule
\multirow{2}{*}{\usecaseb} & \notationa &\cellcolor{palepink}  6.92 & \cellcolor{palepink} 11.34 & \cellcolor{palepink} 9.09 & 6.80 & 10.97 & 10.15 & 6.74 & 15.87 & 7.81 & 9.40 & 3.17 & 4.92 & 11.79 & 11.46 & 10.12 & 8.97 & 10.31 & 10.89 & 5.98 & 7.92 & 9.03 \\  
& \notationc & \cellcolor{palepink}  \textbf{56.28} & \cellcolor{palepink} \textbf{41.94} & \cellcolor{palepink} \textbf{47.93} & \textbf{29.98} & \textbf{29.79} & \textbf{30.55} & \textbf{30.58} & \textbf{54.03} & \textbf{29.27} & \textbf{30.16} & \textbf{51.19} & \textbf{28.21} & \textbf{30.58} & \textbf{30.28} & \textbf{29.63} & \textbf{29.84} & \textbf{30.64} & \textbf{18.07} & \textbf{29.18} & \textbf{25.82} & \textbf{34.20} \\
\midrule
\multirow{2}{*}{\usecasec} & \notationa  &   1.69 & 1.81 &  1.82 & 1.94 & 1.98 & 1.96 & 2.01 & 2.82 & 1.99 & 1.98 & 1.92 & 1.92 & 1.94 & 1.97 & 1.95 & 1.99 & 1.89 & 1.86 & 2.00 & 1.99 & 1.97   \\
& \notationc &  \textbf{38.56} & \textbf{28.00} &  \textbf{43.82} & \textbf{22.98} & \textbf{43.00} & \textbf{23.71} & \textbf{5.73}  & \textbf{22.61} & \textbf{10.65} & \textbf{32.07} & \textbf{20.05} & \textbf{2.13}  & \textbf{44.03} & \textbf{44.61} & \textbf{22.19} & \textbf{20.13} & \textbf{16.52} & \textbf{5.24}  & \textbf{16.83} & \textbf{5.07}  & \textbf{23.40}  \\

\bottomrule
\end{tabular}}
\caption{Results of \notationa on test data of \usecasea, \usecaseb and \usecasec in 20 languages. Test data of \usecasea and \usecaseb in the three languages highlight in pink color are built with MTPE data and the rest are built with MT data.}
\label{tb:global_breakdown}
\end{table*}

We observe that there are a few languages like Thai and Vietnamese have low results than other languages. Through investigation, we found that it was caused by failing to predict the tone marks in most of cases. For example, the model may predict  \textit{“nha khach”} for hotel type while \textit{“{\fontencoding{T5} nhà khách}”} is the ground truth. We may explore options for post-processing or other models to improve the performance on these languages upon the release of the data.